\documentclass[lettersize,journal]{IEEEtran}

\usepackage{amsmath} 
\usepackage{amsmath,amsfonts}
\usepackage{algorithmicx}
\usepackage{algorithm}
\usepackage{array}
\usepackage[caption=false,font=normalsize,labelfont=sf,textfont=sf]{subfig}
\usepackage{textcomp}
\usepackage{stfloats}
\usepackage{url}
\usepackage{amsmath}
\usepackage{amssymb}
\usepackage{color}
\usepackage{algpseudocode}
\usepackage{xcolor}
\usepackage{makecell}
\usepackage{verbatim}
\usepackage{graphicx}
\usepackage{cite}
\usepackage{booktabs}
\usepackage{multirow}
\usepackage{amssymb}
\usepackage[colorlinks,
            linkcolor=blue,
            anchorcolor=blue,
            citecolor=blue]{hyperref}

\begin{document}

\title{Diffusion-Based Restoration for Multi-Modal \\ 3D Object Detection in Adverse Weather}

\author{Zhijian He, 
Feifei Liu,
Yuwei Li,
Zhanpeng Luo,
Jintao Cheng,
Xieyuanli Chen,~\IEEEmembership{Member,~IEEE,}
Xiaoyu Tang$^{*}$,~\IEEEmembership{Member,~IEEE}
\thanks{$^{*}$Corresponding author.}
\thanks{Zhijian He is with the School of Xingzhi College, South China Normal University, Shanwei 516600, China, and the College of Big Data and Internet, Shenzhen Technology University, Shenzhen 518000, China. (e-mail: {\tt\small hezhijian@sztu.edu.cn})}
\thanks{Feifei Liu, Yuwei Li and Zhanpeng Luo are with the School of Data Science and Engineering, Xingzhi College, South China Normal University, Shanwei, 516600, China. (e-mail: {\tt\small \{20238331056, 202481313735, 202481313633\}@m.scnu.edu.cn})}
\thanks{Jintao Cheng is with the Department of Electronic and Computer Engineering, Hong Kong University of Science and Technology, Hong Kong, China. (e-mail: {\tt\small jchengau@connect.ust.hk})}
\thanks{Xieyuanli Chen is with the College of Intelligence Science and Technology, National University of Defense Technology, Changsha, China. (e-mail: {\tt\small xieyuanli.chen@nudt.edu.cn})}
\thanks{Xiaoyu Tang is with the School of Data Science and Engineering, Xingzhi College, South China Normal University, Shanwei, 516600, China. (e-mail: {\tt\small tangxy@scnu.edu.cn})}}

\maketitle

\pagestyle{empty}  
\thispagestyle{empty} 

\begin{abstract}
Multi-modal 3D object detection is important for reliable perception in robotics and autonomous driving. However, its effectiveness remains limited under adverse weather conditions due to weather-induced distortions and misalignment between different data modalities. In this work, we propose DiffFusion, a novel framework designed to enhance robustness in challenging weather through diffusion-based restoration and adaptive cross-modal fusion.
Our key insight is that diffusion models possess strong capabilities for denoising and generating data that can adapt to various weather conditions. Building on this, DiffFusion introduces Diffusion-IR restoring images degraded by weather effects and Point Cloud Restoration (PCR) compensating for corrupted LiDAR data using image object cues. 
To tackle misalignments between two modalities, we develop Bidirectional Adaptive Fusion and Alignment Module (BAFAM). It enables dynamic multi-modal fusion and bidirectional bird's-eye view (BEV) alignment to maintain consistent spatial correspondence.
Extensive experiments on three public datasets show that DiffFusion achieves state-of-the-art robustness under adverse weather while preserving strong clean-data performance. Zero-shot results on the real-world DENSE dataset further validate its generalization.
The implementation of our DiffFusion will be released as open-source.

\end{abstract}

\section{INTRODUCTION}

\IEEEPARstart{P}{recise} 3D object detection remains a key challenge for ensuring the safety of autonomous mobile systems, including robots and self-driving vehicles, particularly in real-world environments affected by rain, fog, glare, and other adverse weather~\cite{dong2023cvpr}. Robust perception under such conditions is essential for reliable deployment, yet existing approaches degrade substantially under weather-induced corruptions.
\begin{figure}[t]
    \centering
    \includegraphics[width=1\linewidth]{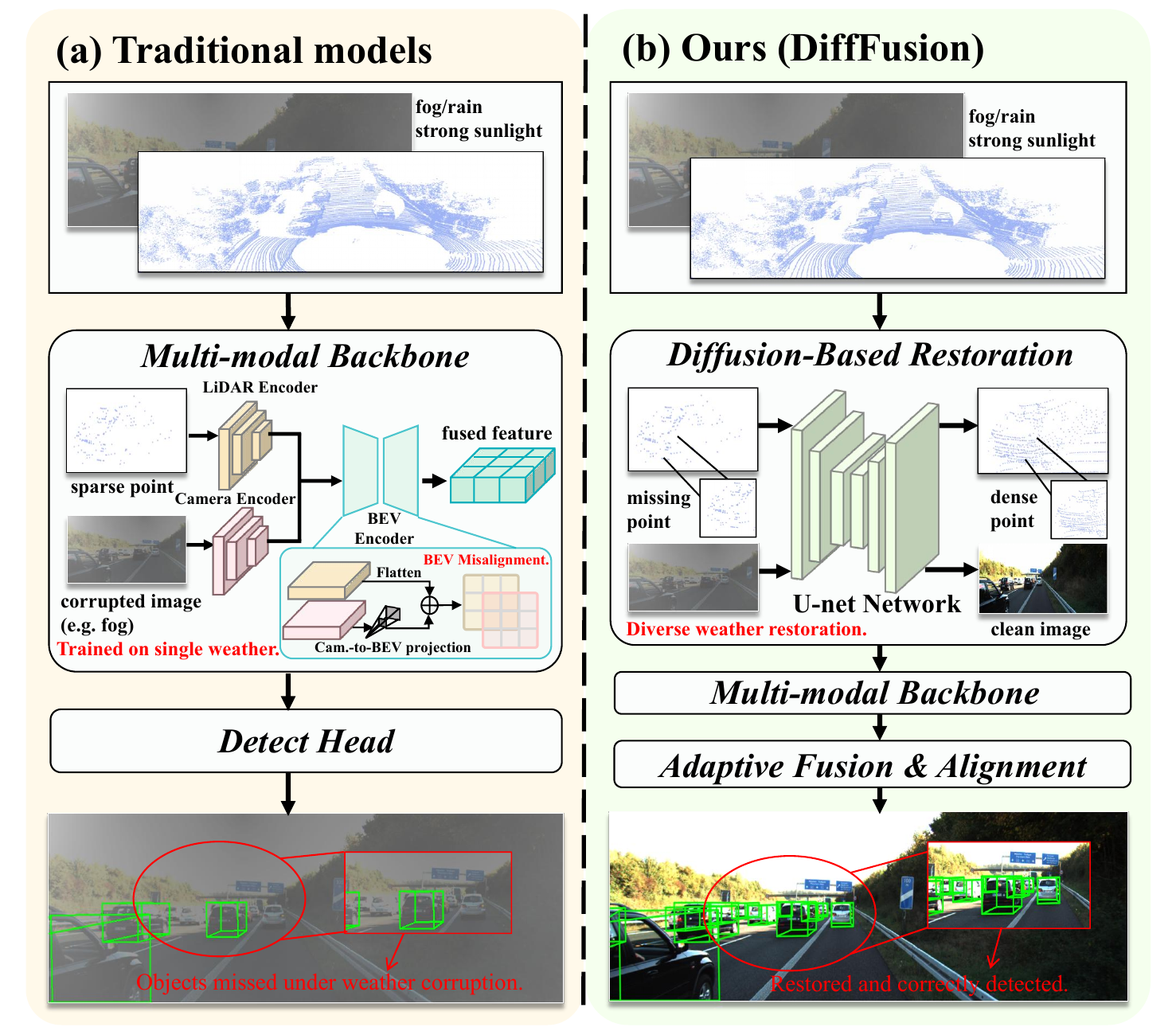}
    \caption{Motivation. (a) Traditional methods are limited to single-weather training and suffer from BEV misalignment when fusing corrupted features, leading to detection failures. (b) DiffFusion employs diffusion-based restoration for diverse weather, with adaptive fusion and alignment enabling reliable 3D detection under adverse weather.}
    \label{fig:core_idea}
    \vspace{-0.5cm}
\end{figure}
Different sensing modalities experience different degradation. For example, images are corrupted by rain streaks, fog, and strong sunlight, while LiDAR measurements are disrupted by airborne particles that scatter and attenuate laser beams~\cite{cheng2024mfmos}. Although multi-modal fusion is widely adopted to enhance robustness, achieving consistent performance under adverse weather remains a critical yet underexplored challenge. Beyond individual sensor degradation, these modality-specific corruptions further disrupt spatial correspondence across sensors, causing inconsistent fusion and in some cases performing even worse than single-modality baselines.

Existing multi-modal fusion methods can be categorized into point-level, proposal-level, and BEV-based paradigms. Among these, BEV-based methods~\cite{BEVfusion, bai2022transfusionrobustlidarcamerafusion, chen2022focalsparseconvolutionalnetworks} have become dominant by projecting both LiDAR and image data into a unified bird's-eye-view space. However, these methods are mainly optimized on clean datasets and employ simplistic fusion strategies such as element-wise summation or concatenation. As a result, they lack explicit mechanisms for handling weather-induced feature corruption and sensor misalignment. Recent efforts have attempted to enhance robustness in adverse conditions. TripleMixer~\cite{triplemixer3dpointcloud} focuses on point-cloud denoising but operates solely on LiDAR and is evaluated on private datasets, limiting generalizability. SAMFusion~\cite{samfusion} fuses four sensing modalities (RGB, Gated, LiDAR, and Radar), but introduces prohibitive computational overhead and still fails to correct cross-sensor alignment errors. 

To overcome these limitations, we propose DiffFusion, a unified framework that systematically addresses weather-induced degradation through diffusion-based multi-modal restoration, as shown in Fig.~\ref{fig:core_idea}. Our key insight is that diffusion models~\cite{ho2020denoisingdiffusionprobabilisticmodels, song2022denoisingdiffusionimplicitmodels} show strong denoising capabilities by learning to invert complex corruption processes rather than relying on handcrafted assumptions. Unlike conventional enhancement techniques designed for specific degradation, diffusion-based restoration demonstrates strong generalization across diverse weather conditions, including rain, fog, and strong sunlight, owing to its robust denoising and generative priors~\cite{rombach2022highresolutionimagesynthesislatent}. Guided by this intuition, we design Diffusion-based Restoration Module, a unified generative restoration module 
that jointly recovers both corrupted images and LiDAR data. Specifically, for the image branch, we employ a conditional denoising 
diffusion network to restore degraded images while preserving detection-relevant 
semantic features. For the LiDAR branch, we leverage object cues extracted from 
restored images to conduct targeted ray compensation via PCR, restoring 
missing points. 
Beyond raw data corruption, existing fusion pipelines often suffer from degraded feature integration due to cross-modal spatial misalignment. To tackle this, we further propose the Bidirectional Adaptive Fusion and Alignment Module (BAFAM), which performs adaptive feature fusion and spatial alignment through two complementary mechanisms.
Extensive experiments on KITTI and KITTI-C demonstrate that DiffFusion achieves state-of-the-art performance under diverse weather conditions, significantly outperforming existing methods while maintaining competitive accuracy on clean datasets. Moreover, zero-shot evaluations on the real-world DENSE dataset further validate the strong generalization capability of our approach.

In summary, our main contributions are threefold:
\begin{enumerate}
\item We introduce DiffFusion, a unified diffusion-based multi-modal restoration framework that systematically recovers weather-degraded sensor data. It comprises a dual-branch design: (i) Diffusion-IR, a conditional denoising diffusion module for image restoration, and (ii) PCR, a LiDAR ray-compensation mechanism guided by object-level cues extracted from restored images.
\item We develop BAFAM, a bidirectional alignment and fusion module that establishes reliable cross-modal correspondence under adverse weather. It integrates: (i) Cross-Attention Adaptive Fusion (CAAF), which performs asymmetric bidirectional cross-attention for dynamically weighted multi-modal fusion, and (ii) Bidirectional BEV Alignment (B2A), which resolves spatial misalignment through cascaded offset learning.
\item Our approach achieves state-of-the-art performance on KITTI and KITTI-C, demonstrating strong robustness across diverse weather conditions. Furthermore, zero-shot results on the real-world DENSE dataset confirm the method’s out-of-distribution generalization capability. We will release our implementation as open-source to support future research. 
\end{enumerate}

\section{Related Work}


\subsection{Multi-Modal 3D Object Detection in Adverse Weather}
Multi-modal 3D object detection integrates LiDAR and camera features to leverage complementary information from both modalities for improved detection performance. Based on the fusion stage in the detection pipeline, previous methods can be classified into three categories: point-level fusion methods~\cite{huang2020epnetenhancingpointfeatures} that fuse features at the raw point cloud level, proposal-level fusion methods~\cite{bai2022transfusionrobustlidarcamerafusion} that combine features at the region proposal stage, and BEV-based methods~\cite{BEVfusion} that perform fusion in a unified bird's-eye-view representation. However, these methods are primarily optimized on clean datasets like nuScenes~\cite{caesar2020nuscenes} and KITTI~\cite{Geiger2012CVPR}, overlooking robustness in adverse weather and lacking explicit alignment mechanisms to handle weather-induced feature misalignment.

Recent works have explored robust detection under adverse weather. 
TripleMixer~\cite{triplemixer3dpointcloud} addresses point cloud denoising but focuses on single-modality LiDAR and is evaluated on private datasets. 
SAMFusion~\cite{samfusion} employs four modalities (RGB, Gated, LiDAR, Radar) but suffers from high computational costs and fails to correct sensor alignment errors. 
Most methods lack systematic solutions 
that jointly address denoising, point corruption patterns, and 
multi-modal misalignment in adverse weather.
\subsection{Generative Models for Image and Point Cloud Restoration}
Generative models, particularly diffusion models, have emerged as powerful tools for restoration tasks by learning complex data distributions. 
Diffusion models have demonstrated remarkable capabilities in image 
restoration and denoising tasks by learning to reverse noise through a 
Markov chain process. Denoising Diffusion Probabilistic Models 
(DDPM)~\cite{ho2020denoisingdiffusionprobabilisticmodels} and Denoising 
Diffusion Implicit Models (DDIM)~\cite{song2022denoisingdiffusionimplicitmodels} 
established the theoretical foundation for iterative denoising. Latent 
diffusion models~\cite{rombach2022highresolutionimagesynthesislatent} 
further reduce computational costs by operating in low-dimensional latent 
space while maintaining high-quality output. Specifically, for adverse 
weather image restoration, Ozdenizci et al.~\cite{adverseweather} 
applied patch-based diffusion models to restore images corrupted by rain, 
snow, and fog.

However, most diffusion-based restoration methods focus on 2D image 
processing and have not been systematically integrated into multi-modal 
3D detection frameworks for autonomous driving. For point cloud restoration, 
various generative approaches have been explored beyond diffusion models. 
Point-cloud upsampling methods~\cite{zeng2021point} have made notable 
progress in producing high-resolution point clouds, while Li et 
al.~\cite{li2022snowing} explored translating point clouds collected under 
adverse weather into clear weather conditions using generative techniques. 
Although these models perform well, they are still not tightly integrated 
with multi-modal 3D detection and do not consider the range-dependent 
corruption patterns specific to LiDAR point clouds in adverse weather, 
where distant points suffer from signal attenuation while nearby points 
are corrupted by weather particle reflections. In contrast, our DiffFusion 
designs a unified framework that jointly processes both image and point 
cloud corruptions while maintaining real-time inference capability.

\begin{figure*}[tbp]
    \centering
    \includegraphics[width=0.95
    \linewidth]{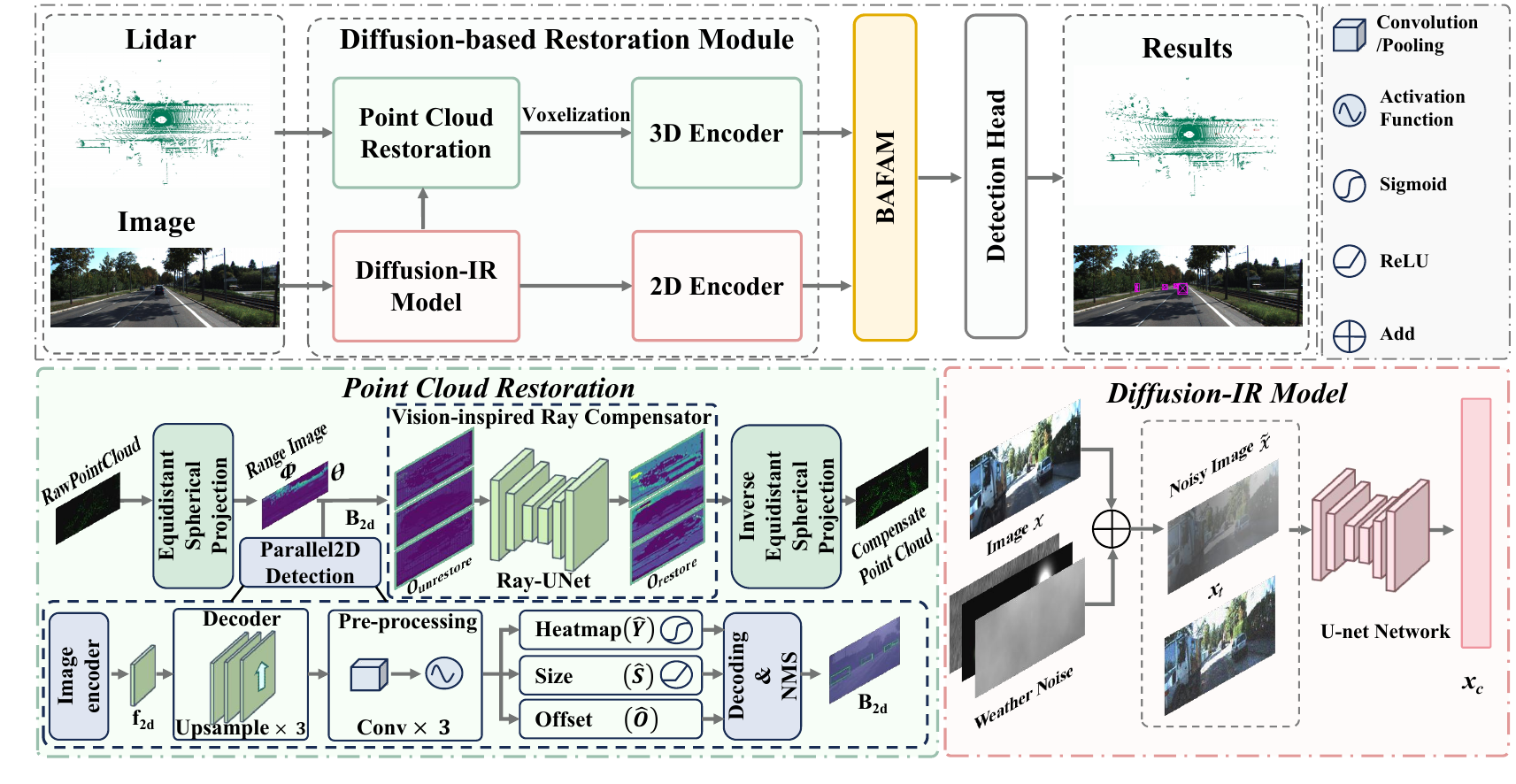}
    \vspace{-0.2cm}
\caption{Overview of the proposed DiffFusion framework. The system first processes weather-corrupted images using Diffusion-IR to extract clean, robust visual features. Then, the image branch provides 2D object detection results to guide the point cloud branch performing depth-aware restoration to recover corrupted LiDAR data. The restored features from both modalities are then processed through the Bidirectional Adaptive Fusion and Alignment Module (BAFAM), which integrates cross-attention fusion with bidirectional spatial alignment to achieve accurate sensor correspondence, producing robust 3D object detection results under adverse weather conditions.}
\vspace{-0.3cm}
\label{fig1}
\end{figure*}

\section{Method}
\subsection{Overall Framework}
In this section, we present DiffFusion, a robust framework that harnesses the generalization capabilities of diffusion models for multi-modal 3D object detection in adverse weather conditions. The overall architecture is depicted in Fig.~\ref{fig1}. 
Given input images and point clouds corrupted by adverse weather conditions, DiffFusion employs the following pipeline: 

(1) \textbf{Diffusion-based Restoration Module.} It systematically addresses weather-induced degradation in both modalities through a dual-branch structure: Diffusion-IR restores weather-degraded images through conditional denoising diffusion to extract robust visual features, while Point Cloud Restoration (PCR) receives guidance from the image branch, utilizing 2D detection bounding boxes combined with depth maps to perform point cloud completion and restoration.

(2) \textbf{Bidirectional Adaptive Fusion and Alignment Module (BAFAM)} integrates the restored features from both modalities through two complementary components: Cross-Attention Adaptive Fusion (CAAF) adaptively combines complementary information from LiDAR and camera modalities, while Bidirectional BEV Alignment (B2A) establishes robust spatial correspondence between modalities.
This unified approach enables DiffFusion to maintain high detection performance across various weather conditions while preserving the complementary advantages of multi-modal sensing.


\subsection{Diffusion-based Restoration Module}

\begin{figure}[tbp]
    \centering
    \includegraphics[width=1.0\linewidth]{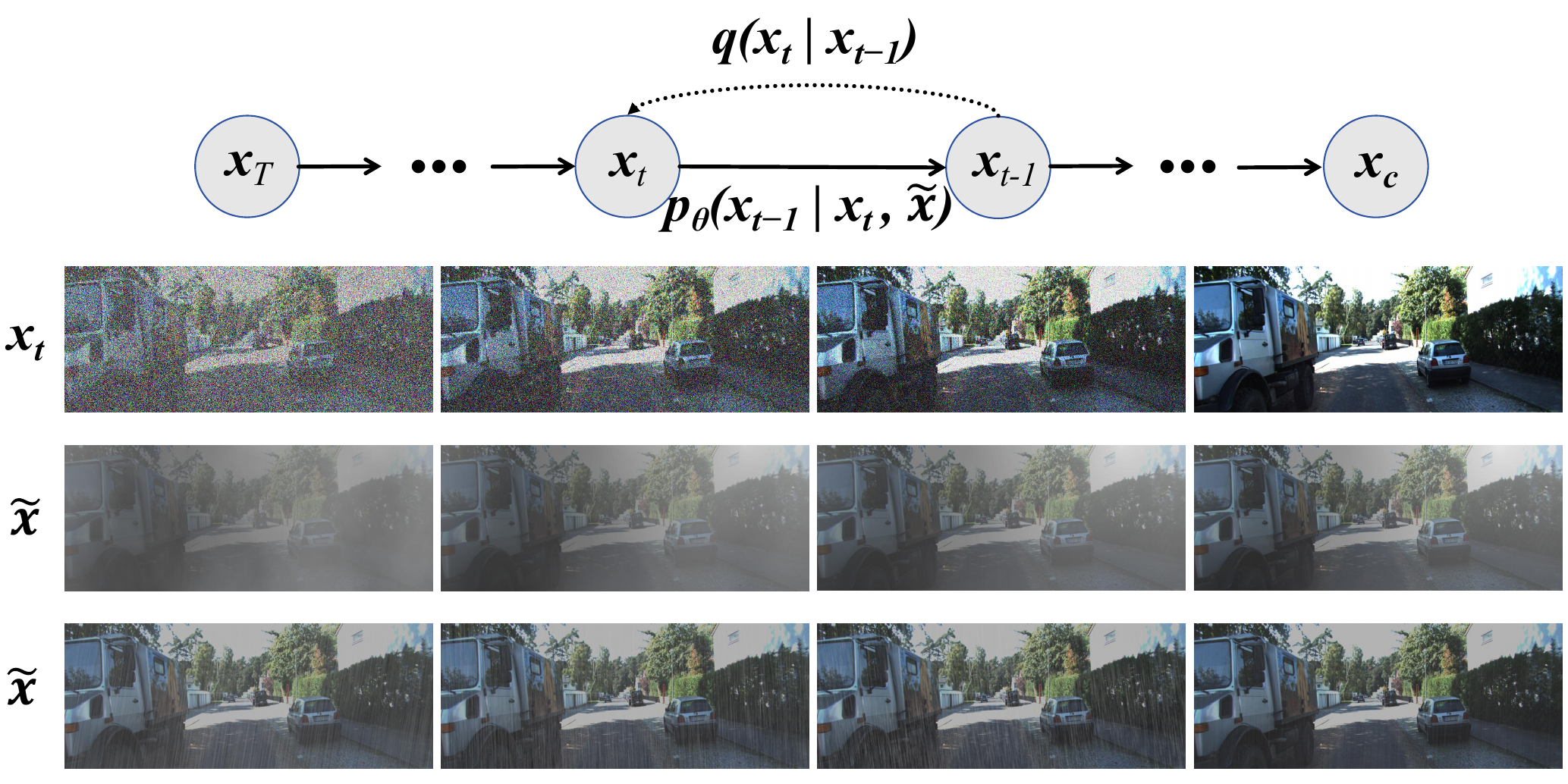}
    \vspace{-0.3cm}
    \caption{Overview of the image branch Diffusion-IR diffusion process. The first row shows the forward diffusion process $q(\mathbf{x}_t|\mathbf{x}_{t-1})$ (dashed line) and the reverse denoising process $p_\theta(\mathbf{x}_{t-1}|\mathbf{x}_t, \tilde{\mathbf{x}})$ (solid line). The third and fourth rows show the restoration process where the diffusion model takes the weather-degraded image $\tilde{\mathbf{x}}$  as additional input to guide the denoising process. Subsequently, the denoised features $\mathbf{F}_{\text{c}}$ are fed into the 2D encoder.}
    \label{fig:diffusion_process}
    \vspace{-0.3cm}
\end{figure}

(1) \textbf{Diffusion-IR.}
We design an image branch with diffusion-based restoration named Diffusion-IR that cascades image restoration with feature extraction. Existing methods directly apply CNNs to corrupted images~\cite{BEVfusion,bai2022transfusionrobustlidarcamerafusion,chen2022focalsparseconvolutionalnetworks}, causing degraded 2D features to propagate errors into the fusion stage. To address this, we employ a conditional denoising diffusion model based on Weather Diffusion~\cite{adverseweather} to restore clean images before feature extraction.

As shown in Fig.~\ref{fig:diffusion_process}, we define a forward process that gradually adds Gaussian noise to clean images over $T$ timesteps following DDPM~\cite{ho2020denoisingdiffusionprobabilisticmodels}:
\begin{equation}
q(\mathbf{x}_t|\mathbf{x}_c) = \mathcal{N}(\mathbf{x}_t; \sqrt{\bar{\alpha}_t}\mathbf{x}_c, (1-\bar{\alpha}_t)\mathbf{I}),
\end{equation}
where $\bar{\alpha}_t = \prod_{i=1}^{t}(1-\beta_i)$ is the variance schedule. This allows direct sampling of noisy image $\mathbf{x}_t$ from clean image $\mathbf{x}_c$:
\begin{equation}
\mathbf{x}_t = \sqrt{\bar{\alpha}_t} \mathbf{x}_c + \sqrt{1 - \bar{\alpha}_t} \boldsymbol{\epsilon}_t, \quad \boldsymbol{\epsilon}_t \sim \mathcal{N}(\mathbf{0}, \mathbf{I}).
\end{equation}

To enable real-time inference, DDIM~\cite{song2022denoisingdiffusionimplicitmodels} reformulates the reverse process as a deterministic mapping using a subsequence of timesteps $\{\tau_1, \tau_2, \ldots, \tau_S\}$ where $S \ll T$:
\begin{equation}
\mathbf{x}_{\tau_{i-1}} = \sqrt{\bar{\alpha}_{\tau_{i-1}}} \cdot \hat{\mathbf{x}}_c + \sqrt{1 - \bar{\alpha}_{\tau_{i-1}}} \cdot \boldsymbol{\epsilon}_\theta(\mathbf{x}_{\tau_i}, \tau_i),
\end{equation}
where $\hat{\mathbf{x}}_c = (\mathbf{x}_{\tau_i} - \sqrt{1-\bar{\alpha}_{\tau_i}} \cdot \boldsymbol{\epsilon}_\theta(\mathbf{x}_{\tau_i}, \tau_i)) / \sqrt{\bar{\alpha}_{\tau_i}}$ is the predicted clean image. This reduces sampling steps from $T=1000$ to $S=10$, achieving significant speedup while preserving restoration quality.

We extend DDIM to a conditional framework where the noise prediction network $\boldsymbol{\epsilon}_\theta$ takes both the noisy latent $\mathbf{x}_t$ and the weather-degraded input $\tilde{\mathbf{x}}$ as conditions. The degraded image is concatenated with $\mathbf{x}_t$ channel-wise, yielding the conditional reverse process:
\begin{equation}
\mathbf{x}_{\tau_{i-1}} = \sqrt{\bar{\alpha}_{\tau_{i-1}}} \cdot \hat{\mathbf{x}}_c(\mathbf{x}_{\tau_i}, \tilde{\mathbf{x}}) + \sqrt{1 - \bar{\alpha}_{\tau_{i-1}}} \cdot \boldsymbol{\epsilon}_\theta(\mathbf{x}_{\tau_i}, \tilde{\mathbf{x}}, \tau_i).
\end{equation}
The network is trained with the following objective:
\begin{equation}
\mathcal{L}_{\text{diff}} = \mathbb{E}_{\mathbf{x}_c, \tilde{\mathbf{x}}, t, \boldsymbol{\epsilon}} \left[ \| \boldsymbol{\epsilon}_t - \boldsymbol{\epsilon}_\theta(\mathbf{x}_t, \tilde{\mathbf{x}}, t) \|^2 \right].
\end{equation}

We pre-train our diffusion model across diverse outdoor driving scenarios to handle diverse weather degradations in a unified manner.
After obtaining the restored image $\mathbf{x}_c$, we employ the image encoder from DiffusionDet~\cite{chen2023diffusiondetdiffusionmodelobject} to extract features. We adopt ResNet-50 as our backbone to process the restored image $\mathbf{x}_c$, producing hierarchical feature maps $\{\mathbf{C}_2, \mathbf{C}_3, \mathbf{C}_4, \mathbf{C}_5\}$ at four stages, where $\mathbf{C}_i \in \mathbb{R}^{B \times C_i \times H/2^i \times W/2^i}$ denotes the feature map with spatial stride $2^i$. To handle multi-scale objects, we employ FPN~\cite{lin2017featurepyramidnetworksobject} to aggregate features across scales. The pyramid features are reduced to 16 channels and upsampled to produce $\mathbf{F}_{\text{c}} \in \mathbb{R}^{B \times 16 \times H/4 \times W/4}$, providing semantically rich representations for multi-modal fusion. 

(1) \textbf{LiDAR Branch with Point Cloud Restoration.}
The LiDAR branch plays a crucial role in perception of outdoor driving scenarios. However, severe point cloud sparsity caused by signal attenuation and adverse weather makes detection more challenging than under normal conditions. To address the problem of heavily missing points in extreme weather, we propose a Point Cloud Restoration module. Unlike existing methods that rely on range images and diffusion models~\cite{nakashima2024lidar} with high computational complexity, we perform point cloud compensation at the object level using range images.

To convert sparse point clouds into a dense representation, we adopt an equidistant spherical projection. Leveraging the pulsed laser characteristics of LiDAR, we map each point in Cartesian coordinates \(p_k = (x_k, y_k, z_k)\) to a pixel position \((u_k^r, v_k^r)\), and uniformly discretize the angular ranges into an \(H \times W\) grid (rows: elevation angle $\theta$, where the vertical observation range is limited from $\theta_{\min}$ to $\theta_{\max}$; columns: azimuth angle $\phi$):

\begin{equation}
    \begin{pmatrix}
        u_k^r \\
        v_k^r
    \end{pmatrix}
    =
    \begin{pmatrix}
        \displaystyle \frac{1}{2}\left(1 - \frac{\phi_k}{\pi}\right) W \\[6pt]
        \displaystyle \left(1 - \frac{\theta_k - \theta_{\min}}{\theta_{\max} - \theta_{\min}}\right) H
    \end{pmatrix}.
\end{equation}

The resulting range image is both dense and compact. It preserves the geometric properties of LiDAR while allowing us to apply standard 2D convolutional networks for efficient feature extraction, thereby improving the performance of subsequent processing stages.

Inspired by the human visual system, which first localizes object regions before inferring their 3D structure, we detect objects on the range image. As illustrated in Fig.~\ref{fig:diffusion_process}, the input image feature \(f_{2d}\) is first obtained by fusing multi-scale FPN features (P2, P3, P4) from the image encoder and reducing their channel dimension to 64. This feature is fed into a CenterNet-inspired~\cite{DBLP:journals/corr/abs-1904-07850} 2D detection head.
The head consists of a three-stage upsampling decoder followed by a three-block convolutional pre-processing module for feature refinement. Subsequently, the refined features are processed by three parallel heads to estimate center probabilities via a Sigmoid heatmap ($\hat{Y}$), predict box dimensions using ReLU activation ($\hat{S}$), and regress center shifts without activation ($\hat{O}$). These predictions are decoded and filtered via Non-Maximum Suppression (NMS) to yield the final 2D bounding boxes \(B_{2d}\). These bounding boxes enable more reliable identification of objects that are otherwise difficult to detect in highly sparse point clouds, thereby facilitating subsequent point cloud restoration. The training loss for 2D detection is defined as:
\begin{equation}
L_{\mathrm{aux}} =  L_{\mathrm{cls}} + L_{\mathrm{reg}}.
\end{equation}

We use the object bounding boxes \(B_{2d}\) predicted by Parallel 2D Detection, map them onto the range image, and extract the regions \(O_{\text{unrestore}}\). A U-Net is then applied to predict and fill in the loss points. Specifically, our compensator adds points in a quantity determined by the range. In the 2D representation, point compensation is implemented via a U-Net-based in-painting scheme. We first upsample the range image within the region so that the original pixels lie on even rows and columns \((2i, 2j)\), while the newly created pixels are zero-padded. Given the existing points, we determine the number of points to add, assigning more points to sparser targets to better capture their geometric structure.

The unrestore patch \(O_{\text{unrestore}}\) is processed with a zero-padding mask \(M\) to produce the restore target region \(O_{\text{restore}}\) in the range image as follows:
\begin{equation}
O_{\text{restore}} = O_{\text{unrestore}} \odot M \;+\; U(O_{\text{unrestore}}) \odot (1 - M),
\end{equation}
where \(U(\cdot)\) denotes the ray compensator that fills in the zero-padded regions, and \(\odot\) is the element-wise product. While preserving the existing points, new points are generated only in the masked regions, using $B_{2d}$ to complete the 3D structure, consistent with the human visual system that infers missing details from object localization.

To optimize the ray generator, we employ a 3D loss function defined on the generated object patch \(O_{\text{gen}}\):
\begin{equation}
\mathcal{L}_{3D} = \mathcal{L}_{G}  + \mathcal{L}_{L1} + \mathcal{L}_{p},
\end{equation}
\begin{equation}
\mathcal{L}_{G} = \left\| O_{gt} G_{x} - O_{gen} G_{x} \right\|_{1} 
+ \left\| O_{gt} G_{y} - O_{gen} G_{y} \right\|_{1},
\end{equation}
where \(G_x\) and \(G_y\) are gradient operators in the horizontal and vertical directions, respectively. The ground-truth (GT) object patch in the clean range image is denoted by \(O_{gt}\). Moreover, \(\mathcal{L}_{p}\) denotes the perceptual loss. This pretrained ray compensator is then used during the detection stage.

\subsection{Bidirectional Adaptive Fusion and Alignment}

Existing BEV-based multi-modal methods either ignore feature misalignment entirely~\cite{BEVfusion} or adopt unidirectional alignment strategies~\cite{song2023graphbev} that assume one modality provides more accurate spatial positioning than the other. This assumption is fundamentally flawed for two reasons: First, neither modality is consistently more reliable. LiDAR suffers from signal attenuation in adverse weather, while cameras degrade under poor illumination. Second, simplistic fusion strategies such as element-wise summation fail to leverage complementary strengths, resulting in suboptimal correspondence and degraded performance on small objects.

To address these limitations, we propose the Bidirectional Adaptive Fusion and Alignment Module (BAFAM). BAFAM consists of two complementary components: Cross-Attention Adaptive Fusion (CAAF) for adaptive multi-modal feature integration, and Bidirectional BEV Alignment (B2A) for robust spatial correspondence.

\subsubsection{Cross-Attention Adaptive Fusion}

Unlike conventional fusion methods that employ simple element-wise summation or concatenation, we design Cross-Attention Adaptive Fusion (CAAF) to adaptively integrate complementary information from both modalities through asymmetric bidirectional cross-attention.

Given camera BEV features $F^{C}_{B}$ and LiDAR BEV features $F^{L}_{B}$, we first apply adaptive average pooling to obtain fixed-length representations, which effectively controls memory consumption while preserving essential spatial information.
We implement a multi-head cross-attention mechanism as:
\begin{equation}
\mathcal{A}(Q, K, V) = \text{softmax}\left(\frac{QK^{T}}{\sqrt{d}}\right)V,
\end{equation}
where $Q$, $K$, $V$ are query, key, and value matrices projected from input features, and $d$ is the scaling factor.

Our CAAF consists of two stages. In the first stage, LiDAR features serve as queries while camera features provide keys and values, enabling LiDAR to selectively attend to relevant visual information:
\begin{equation}
\hat{F}^{L}_{B} = F^{L}_{B} + \mathcal{A}(F^{L}_{B}, F^{C}_{B}, F^{C}_{B}).
\end{equation}

In the second stage, camera features serve as queries while the corrected LiDAR features provide keys and values:
\begin{equation}
\hat{F}^{C}_{B} = F^{C}_{B} + \mathcal{A}(F^{C}_{B}, \hat{F}^{L}_{B}, \hat{F}^{L}_{B}).
\end{equation}

The final fused feature is obtained by averaging both representations, enhanced through an MLP with residual connections, and combined with dual residual connections from both input modalities:
\begin{equation}
\bar{F}^{L}_{B} = \text{MLP}\left(\frac{\hat{F}^{C}_{B} + \hat{F}^{L}_{B}}{2}\right) + F^{C}_{B} + F^{L}_{B}.
\end{equation}

This asymmetric bidirectional design allows LiDAR features to be corrected using camera information, and then enables camera features to leverage the corrected LiDAR for comprehensive fusion.

\subsubsection{Bidirectional BEV Alignment}

To further address spatial misalignment between modalities, we design Bidirectional BEV Alignment (B2A) that performs cascaded offset learning for robust feature correspondence. B2A takes the fused LiDAR feature $\bar{F}^{L}_{B}$ from CAAF and the original camera BEV feature $F^{C}_{B}$ as inputs.

We first generate the supervision target by concatenating clean LiDAR and camera BEV features to obtain $F^{LC}_{B} \in \mathbb{R}^{B_S \times (C_{C} + C_{L}) \times H_{B} \times W_{B}}$. Subsequently, $F^{LC}_{B}$ undergoes convolution operations through a CBR-module, resulting in supervision features $F^{S}_{B} \in \mathbb{R}^{B_S \times C_{L} \times H_{B} \times W_{B}}$, representing the optimal fusion without misalignment.

In the first stage, we introduce random spatial shifts to $\bar{F}^{L}_{B}$ to simulate misalignment scenarios, obtaining shifted features $\bar{F}^{L\prime}_{B}$. We then concatenate $F^{C}_{B}$ and $\bar{F}^{L\prime}_{B}$ as input to an offset prediction module implemented with CBR-modules. This module learns to predict spatial offsets $F^{O_1} \in \mathbb{R}^{B_S \times 2 \times H_{B} \times W_{B}}$, where the two channels correspond to offset coordinates $(u, v)$. Subsequently, the predicted offsets $F^{O_1}$ are normalized and applied through grid sampling on $F^{C}_{B}$ to generate spatial weights $F^{W_1} \in \mathbb{R}^{B_S \times C_{C} \times H_{B} \times W_{B}}$. Then, $F^{W_1}$ is element-wise multiplied with $F^{C}_{B}$ to dynamically adjust the features, followed by convolution operations to produce the aligned camera BEV feature $\tilde{F}^{C}_{B}$.

In the second stage, we introduce random offset noise to $\tilde{F}^{C}_{B}$ to simulate residual misalignment, obtaining $\tilde{F}^{C\prime}_{B}$. The shifted aligned camera BEV is then concatenated with $\bar{F}^{L}_{B}$ as input to another offset prediction module. Following the same procedure, we predict offsets $F^{O_2} \in \mathbb{R}^{B_S \times 2 \times H_{B} \times W_{B}}$, perform grid sampling on $\bar{F}^{L}_{B}$ to obtain spatial weights $F^{W_2} \in \mathbb{R}^{B_S \times C_{L} \times H_{B} \times W_{B}}$, and apply convolution to generate the final aligned LiDAR BEV feature $\tilde{F}^{L}_{B} \in \mathbb{R}^{B_S \times C_{L} \times H_{B} \times W_{B}}$.

The alignment losses are formulated as:
\begin{equation}
\mathcal{L}_{a} = \lambda_1 \mathcal{L}_{1} + \lambda_2 \mathcal{L}_{2}
\end{equation}
\begin{equation}
\mathcal{L}_{1} = \frac{1}{N_{B}} \sum_{i=1}^{N_{B}} \left(F^{S}_{B,i} - \text{Conv}(\tilde{F}^{C\prime}_{B}, \bar{F}^{L}_{B})_i\right)^2
\end{equation}
\begin{equation}
\mathcal{L}_{2} = \frac{1}{N_{B}} \sum_{i=1}^{N_{B}} \left(F^{S}_{B,i} - \tilde{F}^{L}_{B,i}\right)^2
\end{equation}
where $\lambda_1=0.3$ and $\lambda_2=0.7$ to emphasize the final alignment. During inference, offset noise is removed and the learned offsets are directly applied. The final aligned feature $\tilde{F}^{L}_{B}$ is fed to the detection head for 3D object detection.

\section{Experiments}
\subsection{Datasets}

We conduct extensive experiments on both clean KITTI and corrupted KITTI-C benchmarks to assess our method's robustness in diverse weather conditions.

\textbf{KITTI.} The KITTI dataset~\cite{Geiger2012CVPR} provides synchronized LiDAR point clouds and front-view camera images captured in urban, rural, and highway scenarios. It consists of 3,712 training samples, 3,769 validation samples, and 7,518 test samples. Following the standard evaluation protocol, we report 3D object detection performance using Average Precision (AP) with 40 recall positions (R40) across three difficulty levels: Easy, Moderate, and Hard. We use the mean AP (mAP) averaged across these three difficulty levels as our primary evaluation metric for the clean benchmark.

\textbf{KITTI-C.} To evaluate model robustness under adverse conditions, we adopt the KITTI-C benchmark established by~\cite{dong2023cvpr}. This benchmark synthesizes 27 types of common corruptions for both LiDAR and camera sensors, systematically evaluating the corruption robustness of 3D object detectors. Each corruption type is generated at five severity levels (from light to severe) to simulate varying degrees of degradation. In our experiments, we focus on three weather-related corruptions that are most relevant to real-world autonomous driving: rain, fog, and strong sunlight. Following the benchmark protocol, we train our model on the clean training set and evaluate on the corrupted validation set. We report the mean AP (mAP) of Moderate difficulty averaged across all five severity levels as our main metric, which provides a comprehensive assessment of robustness under varying weather degradation intensities.

\textbf{DENSE (Seeing Through Fog).} The DENSE dataset~\cite{STF} was collected through over 10,000 km of driving in northern Europe and provides real fog samples with visibility ranging from 20m to 100m. The dataset contains approximately 12,000 samples along with 1,500 samples captured under controlled conditions in a fog chamber with 3D bounding box annotations. In our experiments, we use 442 fog samples from DENSE as an additional validation set to demonstrate the generalization capability of our approach to real-world fog scenarios. An IoU threshold of 0.5 was used for evaluation.

\subsection{Experimental Setup}

\textbf{Implementation Details.} All experiments are conducted on NVIDIA RTX4090 GPUs with 24GB memory. We implement our DiffFusion framework based on the Focals Conv~\cite{chen2022focalsparseconvolutionalnetworks} baseline architecture. The input voxel size is set to (0.05m, 0.05m, 0.1m) for both LiDAR and image features. For the car class, we use anchor sizes of [3.9, 1.6, 1.56] and anchor rotations at [0, 1.57] radians. We adopt the same data augmentation strategy as Focals Conv to ensure fair comparison.

\begin{figure*}[tbp]
    \centering
    \includegraphics[width=1.0\linewidth]{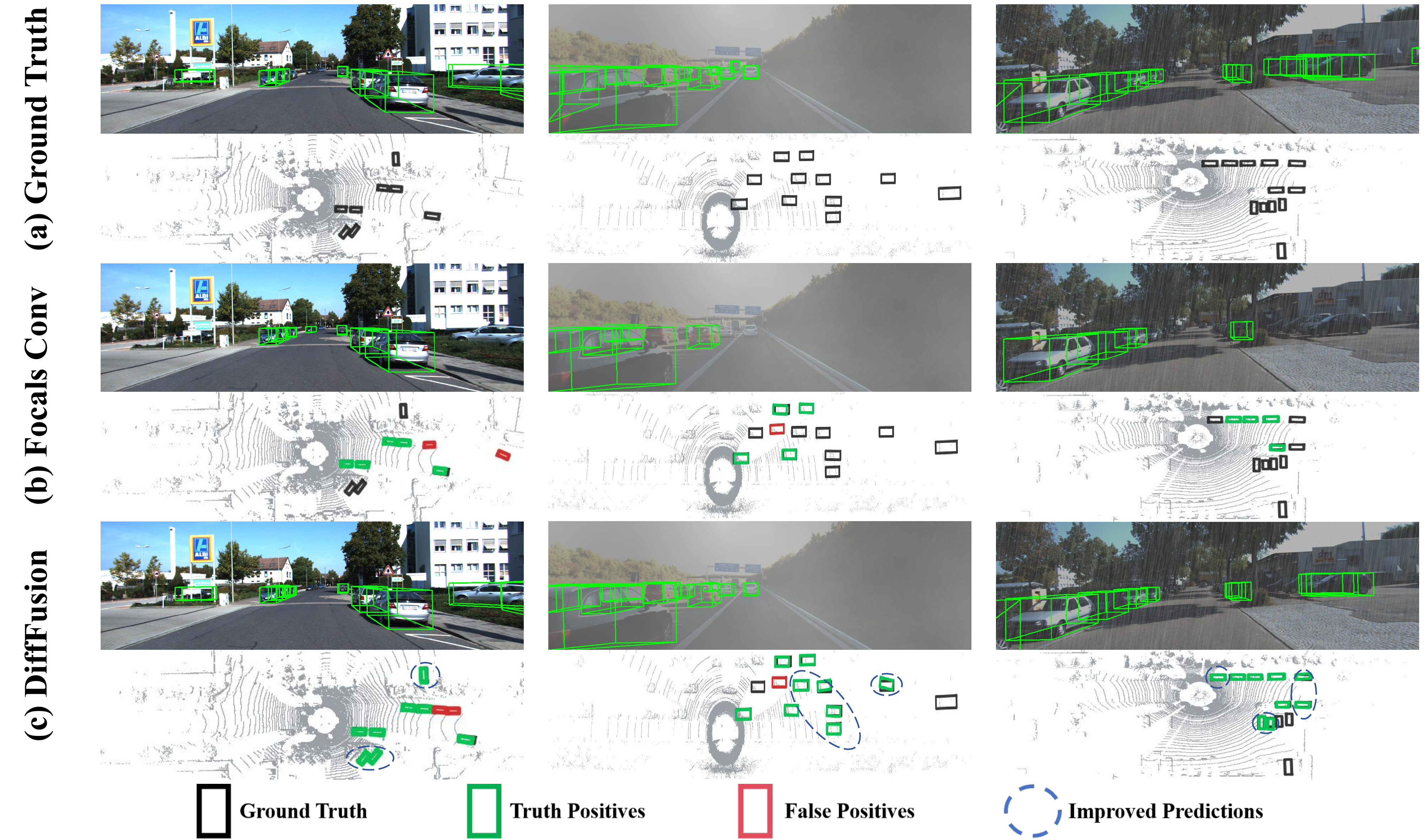}
    \vspace{-0.2cm}
\caption{Visualization Results of Focals Conv and our DiffFusion in KITTI-C dataset. We use boxes in red to represent false positives, green boxes for truth positives, and black for the ground truth. We use blue dashed ovals to highlight the pronounced improvements in predictions.}
\label{visual}
    \vspace{-0.4cm}
\end{figure*}

\begin{table}[tbp]
\setlength{\abovecaptionskip}{0.2cm}
\setlength{\belowcaptionskip}{-0.1cm}
\centering
\caption{Comparison with SOTA methods on KITTI validation set for car class with AP of $R_{40}$. 'Mod.' denotes Moderate.}
\label{tab:kitti_val_car}
\setlength{\tabcolsep}{2.5pt}
\footnotesize
\begin{tabular}{@{}c|c|cccc@{}}
\toprule
\textbf{Method} & \textbf{Type} & \textbf{mAP} & \textbf{Easy} & \textbf{Mod.} & \textbf{Hard} \\
\midrule
SMOKE~\cite{liu2020smokesinglestagemonocular3d} & \multirow{3}{*}{Camera-Only} & 7.69 & 10.42 & 7.09 & 5.57 \\
PGD~\cite{wang2021probabilisticgeometricdepthdetecting} & & 9.01 & 12.72 & 8.10 & 6.20 \\
ImVoxelNet~\cite{rukhovich2021imvoxelnetimagevoxelsprojection} & & 12.85 & 17.85 & 11.49 & 9.20 \\
\cmidrule(lr){1-6}
PointPillars~\cite{8954311} & \multirow{6}{*}{LiDAR-Only} & 80.45 & 87.75 & 78.41 & 75.19 \\
3DSSD~\cite{yang20203dssdpointbased3dsingle} & & 83.11 & 91.07 & 80.03 & 78.23 \\
PointRCNN~\cite{shi2019pointrcnn3dobjectproposal} & & 83.43 & 91.65 & 80.57 & 78.06 \\
SECOND~\cite{DBLP:journals/sensors/YanML18} & & 83.56 & 90.53 & 81.59 & 78.57 \\
Part-A$^2$~\cite{shi2020pointsparts3dobject} & & 84.78 & 91.68 & 82.45 & 80.22 \\
PV-RCNN~\cite{shi2022pvrcnnpointvoxelfeatureset} & & 85.66 & 92.10 & 84.39 & 82.49 \\
\cmidrule(lr){1-6}
EPNet~\cite{huang2020epnetenhancingpointfeatures} & \multirow{4.5}{*}{Multi-modal} & 85.06 & \underline{92.29} & 82.72 & 80.16 \\
Focals Conv-F~\cite{chen2022focalsparseconvolutionalnetworks} & & \underline{87.08} & 92.00 & \textbf{85.88} & \underline{83.36} \\
LoGoNet~\cite{li2023logonetaccurate3dobject} & & \textbf{87.13} & 92.04 & 85.04 & \textbf{84.31} \\
\cmidrule(lr){1-1}\cmidrule(lr){3-6}
\textbf{DiffFusion} & & 86.80 & \textbf{92.42} & \underline{85.14} & 82.85 \\
\bottomrule
\end{tabular}
\vspace{-0.3cm}
\end{table}

\subsection{Comparison with State-of-the-Art Methods}

\textbf{Results on clean benchmark.} Tab.~\ref{tab:kitti_val_car} presents the comparison with state-of-the-art methods on the KITTI validation set for the car class. Our DiffFusion achieves SOTA performance on clean data, demonstrating that our approach designed for adverse weather conditions maintains high accuracy in normal scenarios.

\begin{table}[tbp]
\setlength{\abovecaptionskip}{0.2cm}  
\setlength{\belowcaptionskip}{-0.1cm} 
\centering
\caption{Comparison with different methods on KITTI-C dataset. 'S.L.' denotes Strong Sunlight.}
\label{tab:kittic}
\setlength{\tabcolsep}{2.5pt} 
\footnotesize 
\begin{tabular}{@{}c|c|c|cccc@{}}
\toprule
\multirow{2}{*}{\textbf{Method}} & \multirow{2}{*}{\textbf{Type}} & \multirow{2}{*}{\textbf{Clean}} & \multicolumn{4}{c}{\textit{Weather}} \\
 &  &  & \textbf{mAP} & \textbf{Rain} & \textbf{Fog} & \textbf{S.L.} \\
\midrule
SMOKE~\cite{liu2020smokesinglestagemonocular3d} & \multirow{3}{*}{Camera-Only} & 7.09 & 5.19 & 3.94 & 5.63 & 6.00 \\
PGD~\cite{wang2021probabilisticgeometricdepthdetecting} & & 8.10 & 3.67 & 3.06 & 0.87 & 7.07 \\
ImVoxelNet~\cite{rukhovich2021imvoxelnetimagevoxelsprojection} & & 11.49 & 4.22 & 1.24 & 1.34 & 10.08 \\
\cmidrule(lr){1-7}
PointPillars~\cite{8954311} & \multirow{6}{*}{LiDAR-Only} & 78.41 & 54.25 & 36.18 & 64.28 & 62.28 \\
3DSSD~\cite{yang20203dssdpointbased3dsingle} & & 80.03 & 32.75 & 26.28 & 45.89 & 26.09 \\
PointRCNN~\cite{shi2019pointrcnn3dobjectproposal} & & 80.57 & 62.06 & 51.27 & 72.14 & 62.78 \\
SECOND~\cite{DBLP:journals/sensors/YanML18} & & 81.59 & 68.32 & 52.55 & 74.10 & 78.32 \\
Part-A$^2$~\cite{shi2020pointsparts3dobject} & & 82.45 & 63.23 & 41.63 & 71.61 & 76.45 \\
PV-RCNN~\cite{shi2022pvrcnnpointvoxelfeatureset} & & 84.39 & 70.32 & 51.58 & 79.47 & 79.91 \\
\cmidrule(lr){1-7}
EPNet~\cite{huang2020epnetenhancingpointfeatures} & \multirow{4}{*}{Multi-modal} & 82.72 & 50.09 & 36.27 & 44.35 & 69.65 \\
Focals Conv-F~\cite{chen2022focalsparseconvolutionalnetworks} & & 85.88 & 55.61 & 41.30 & 44.55 & 80.97 \\
LoGoNet~\cite{li2023logonetaccurate3dobject} & & 85.04 & 66.29 & 55.80 & 67.53 & 75.54 \\
\cmidrule(lr){1-1}\cmidrule(lr){3-7}
\textbf{DiffFusion} & & \textbf{85.14} & \textbf{71.98} & \textbf{60.31} & \textbf{73.37} & \textbf{82.27} \\ 
\bottomrule
\end{tabular}
\end{table}

\begin{table}[tbp]
\setlength{\abovecaptionskip}{0.2cm}
\setlength{\belowcaptionskip}{-0.1cm}
\centering
\caption{Comparison with different methods on DENSE dataset (Fog) for car class with AP of $R_{40}$.}
\label{tab:dense_fog_car}
\setlength{\tabcolsep}{2.5pt}
\footnotesize
\begin{tabular}{@{}c|c|ccc@{}}
\toprule
\textbf{Method} & \textbf{Type} & \textbf{Easy} & \textbf{Mod.} & \textbf{Hard} \\
\midrule
PointPillars~\cite{8954311} & \multirow{6}{*}{LiDAR-Only} & 25.62 & 23.94 & 21.67 \\
SECOND~\cite{DBLP:journals/sensors/YanML18} & & 32.37 & 30.46 & 27.76 \\
PointRCNN~\cite{shi2019pointrcnn3dobjectproposal} & & 32.00 & 28.60 & 26.34 \\
Part-A$^2$~\cite{shi2020pointsparts3dobject} & & 23.03 & 21.30 & 19.06 \\
PV-RCNN~\cite{shi2022pvrcnnpointvoxelfeatureset} &  & 14.16 & 13.75 & 12.42 \\
Voxel R-CNN~\cite{deng2021voxelrcnnhighperformance} &  & 13.92 & 13.21 & 12.17 \\
\cmidrule(lr){1-5}
Focals Conv-F~\cite{chen2022focalsparseconvolutionalnetworks} & \multirow{2.5}{*}{Multi-modal}  & 4.17 & 4.25 & 4.29 \\
\cmidrule(lr){1-1}\cmidrule(lr){3-5}
\textbf{DiffFusion} &  & \textbf{34.39} & \textbf{31.89} & \textbf{28.76} \\
\bottomrule
\end{tabular}
\vspace{-0.5cm}
\end{table}

\begin{figure}[t]
    \centering
    \includegraphics[width=1.0\linewidth]{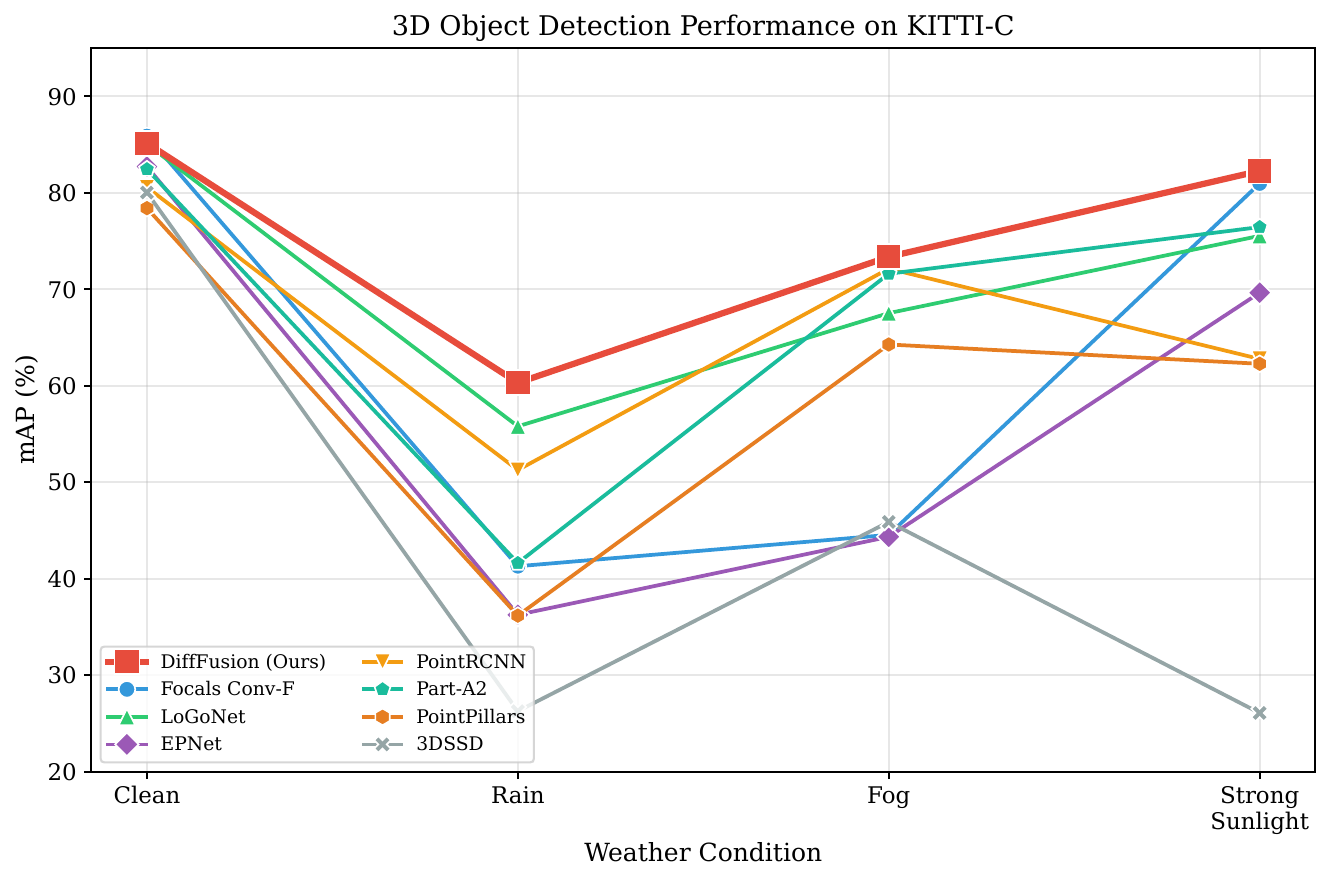}
    \caption{Performance comparison on KITTI-C benchmark across different weather conditions. Our DiffFusion consistently outperforms both LiDAR-only and multi-modal methods under adverse weather while maintaining competitive accuracy on clean data.}
    \label{fig:weather_comparison}
\end{figure}

\textbf{Results on noisy benchmark.} 
Tab.~\ref{tab:kittic} demonstrates the robustness evaluation on KITTI-C dataset under three adverse weather corruptions: fog, rain, and strong sunlight. For fair comparison, we evaluate our method against the same state-of-the-art models as in Tab.~\ref{tab:kitti_val_car}. As shown in Tab.~\ref{tab:kittic}, All detection models across different modalities exhibit varying degrees of performance degradation under adverse weather conditions. Camera-only methods suffer from severe performance drops under rain corruption, significantly lower than LiDAR-only methods. Notably, rain corruption causes more severe degradation than fog across all methods, indicating that rain induces more substantial damage to point clouds. For instance, the LiDAR-only methods like PV-RCNN experiences a 38.9\% performance drop under rain, while the multi-modal methods like LoGoNet shows a 35.2\% degradation. Although multi-modal methods demonstrate slightly better robustness than single-modal approaches, they still suffer from significant performance loss, which reveals that noisy image branches can negatively impact detection performance in multi-modal frameworks. This comparison highlights the necessity of improving both LiDAR and camera branches respectively. Overall, our DiffFusion achieves SOTA performance on the KITTI-C dataset under all corrupted conditions, demonstrating superior robustness and generalization capability. We emphasize not only achieving high accuracy but also ensuring the robustness and generalization of the detection method.
As shown in Fig.~\ref{visual}, our DiffFusion outperforms the baseline Focals Conv-F and achieves balanced performance across all corruption types. This is because our BAFAM and Cross-Fusion strategy dynamically adjusts the contribution of each modality based on weather conditions, effectively combining the complementary strengths of both LiDAR and camera data.

To validate the out-of-distribution generalization capability, we evaluate on the DENSE dataset~\cite{STF} by directly applying models trained on KITTI without fine-tuning. As shown in Tab.~\ref{tab:dense_fog_car}, all methods exhibit performance degradation due to the domain gap between synthetic and real-world fog. Notably, the multi-modal baseline Focals Conv-F suffers a dramatic performance drop (from 85.88\% on KITTI to only 4.25\% on DENSE), indicating that naive fusion strategies fail to generalize when facing real-world fog patterns. In contrast, our DiffFusion achieves the best performance, demonstrating that our diffusion-based restoration module effectively handles the domain shift between synthetic corruptions and real-world fog degradation.

We further visualize the performance trends of KITTI-C in Fig.~\ref{fig:weather_comparison}. The results clearly demonstrate that DiffFusion consistently outperforms both LiDAR-only methods (e.g., PointRCNN, Part-A$^2$) and multi-modal methods (e.g., Focals Conv-F, LoGoNet) across all adverse weather conditions, validating the effectiveness of our unified restoration and fusion framework.

\begin{table}[t]
\centering
\caption{Ablation experiments with proposed modules on KITTI-C validation set. We report Moderate mAP for car class under different weather corruptions and inference speed (FPS) on NVIDIA RTX4090 GPUs.}
\label{tab:ablation}
\setlength{\tabcolsep}{4pt}
\footnotesize
\begin{tabular}{@{}c|cc|ccc|c@{}}
\toprule
\multirow{2}{*}{Method} & \multicolumn{2}{c|}{Component} & \multicolumn{3}{c|}{Weather Corruption} & \multirow{2}{*}{FPS} \\
\cmidrule(lr){2-3} \cmidrule(lr){4-6}
 & DBRM & BAFAM & Rain & Fog & S.L. & \\
\midrule
a)  & - & - & 41.30 & 44.55 & 80.97 & 6.2 \\
b)  & \checkmark & - & 59.24 & 69.92 & 81.54 & 4.4 \\
c)  & \checkmark & \checkmark & 60.31 & 73.37 & 82.27 & 4.3 \\
\bottomrule
\end{tabular}
\vspace{0.1cm}
\\
\footnotesize{\textit{DBRM}: Diffusion-Based Restoration Module, \textit{BAFAM}: Bidirectional Adaptive Fusion and Alignment Module}
\vspace{-0.5cm}
\end{table}
\subsection{Ablation Study}
Tab.~\ref{tab:ablation} presents a systematic ablation study evaluating each proposed component using Focals Conv~\cite{chen2022focalsparseconvolutionalnetworks} as the baseline. The baseline demonstrates significant performance degradation under adverse weather conditions, achieving only (41.30\%, 44.55\%, 80.97\%) mAP under rain, fog, and strong sunlight respectively. Leveraging the strong denoising capabilities of diffusion models, DBRM yields substantial improvements from baseline to (59.24\%, 69.92\%, 81.54\%), demonstrating that explicit weather restoration effectively removes corruption patterns from both image and LiDAR modalities. Subsequently, BAFAM achieves even higher performance to (59.01\%, 72.14\%, 82.65\%), highlighting the effectiveness of adaptive fusion and bidirectional alignment in establishing robust cross-modal correspondence. This further validates the substantial contributions of each module within our DiffFusion framework in addressing weather-induced degradation in autonomous driving scenarios.

\section{CONCLUSIONS}
In this paper, we propose DiffFusion, a robust framework that leverages diffusion models to enhance multi-modal 3D object detection in adverse weather conditions. By integrating diffusion-based restoration with adaptive cross-modal fusion and alignment, our method effectively addresses weather-induced sensor degradation and spatial misalignment between modalities. Extensive experiments on KITTI-C and DENSE datasets demonstrate that DiffFusion achieves state-of-the-art performance across various weather conditions while maintaining competitive results on clean datasets. The strong out-of-distribution generalization to real-world fog scenarios further validates the practical applicability of our framework. Future work will explore real-time optimization and extension to additional sensing modalities.

\bibliographystyle{ieeetr}
\bibliography{root.bib}
\end{document}